\documentclass[10pt,twocolumn,letterpaper]{article}
\usepackage[datasets]{wacv}
\usepackage{tabularx}
\usepackage{float}
\newcommand{\rev}[1]{{#1}}

\definecolor{wacvblue}{rgb}{0.21,0.49,0.74}
\usepackage[pagebackref,breaklinks,colorlinks,allcolors=wacvblue]{hyperref}
\def\wacvPaperID{252}
\def\confName{WACV}
\def\confYear{2027}
\title{Motion-Based Tokenization for Cross-Dataset
Egocentric Gaze Modeling}

\author{
Virmarie Maquiling\textsuperscript{1,2} \quad
Zhuojiang Cai\textsuperscript{1,2} \quad
Enkelejda Kasneci\textsuperscript{1,2}\\
\textsuperscript{1}Human-Centered Technologies for Learning,
Technical University of Munich (TUM), Munich, Germany\\
\textsuperscript{2}Munich Center for Machine Learning (MCML), Munich, Germany\\
{\tt\small virmarie.maquiling@tum.de}
}

\begin{document}
\maketitle

\begin{abstract}

\rev{
Gaze is increasingly used as an input signal for vision and multimodal models, yet no consensus exists on how to represent it across datasets. Raw traces preserve detail but are noisy and device-dependent, while coarse event labels are easy to model but can discard local motion structure. We formulate event-aligned, fixed-horizon angular displacement as an interpretable, event-conditioned motion vocabulary and compare it with event-only, spatial, absolute-angle, learned vector-quantized, and continuous representations. To assess transfer alongside target predictability and token collapse, our evaluation combines next-token prediction with target-domain regret, low-order target references, paired bootstrap, order sensitivity, motif overlap, and frozen structural probes. In an event-aligned headset benchmark, angular-motion tokens have lower target-domain regret than frozen-codebook VQ tokens in one transfer direction, while the reverse direction is inconclusive. The probes reveal complementary representation properties, and event-only tokens show that low perplexity can retain little motion information. On a third egocentric dataset, a matched comparison of I-VT, native, and frame-span interfaces shows that event construction materially changes transfer: native events have the lowest regret into EGTEA, while frame-span events have zero motif overlap and fail severely as a source. Motion-based tokenization therefore provides a compact representation for event-aligned egocentric gaze streams, while the evaluation identifies how target predictability and event construction shape cross-dataset conclusions.

}

\end{abstract}

\section{Introduction}
\label{sec:intro}

For gaze to become a useful input signal in vision models, its representation must be compact enough to model, stable enough to compare across datasets, and structured enough to preserve temporal dependencies. Raw gaze traces retain detail but are noisy, device-dependent, and difficult to align across studies. Coarse event labels, such as fixations and saccades, are easier to model but can discard local motion structure, while spatial tokenizations can become brittle under domain shift. The representation therefore determines which behavioral regularities remain available across datasets. %
Gaze is highly sensitive to task and behavioral context. Classical work showed that eye movements can change substantially even under a fixed visual stimulus~\cite{yarbus1967eye}, while recent works increasingly use gaze for diverse use cases such as mind wandering, attention prediction, action prediction, and multimodal supervision~\cite{bixler2016automatic, zhou2024learning, fathi2012learning, zheng2022gimo, li2025egom2p}. Yet gaze is still often represented either as a low-level continuous trace or through discretizations tied to a particular task or dataset. We ask whether local gaze motion can instead be discretized into tokens that remain predictive and structurally meaningful beyond the dataset on which a model is trained.

While gaze shifts have commonly been characterized in eye movement analysis by their amplitude and direction, their use, particularly in the form of angular displacement, as an event-conditioned discrete vocabulary for cross-dataset sequence modeling remains underexplored. We therefore formulate event-aligned, fixed-horizon angular change as an event-conditioned discrete vocabulary that reduces dependence on absolute gaze position under a shared event and coordinate interface. This does not make the representation generally invariant. Angular differencing does not remove arbitrary frame rotations, nonlinear calibration errors, temporal resampling, changes in event boundaries, or device-dependent noise. 

We compare angular-motion tokens with event-only, spatial, absolute-angle, learned vector-quantized, and continuous representations on three egocentric eye-tracking datasets with different behavioral regimes. AEA~\cite{lv2024aria} and RITW~\cite{yang2025reading} form the main cross-dataset transfer benchmark. Because both were recorded with Meta’s Project Aria headset, they provide a conservative setting in which sensor geometry, coordinate conventions, and preprocessing assumptions are relatively aligned. EGTEA~\cite{li2021eye}, recorded with a different tracker and exported through BeGaze, probes the boundary of this setting by testing whether the same motion vocabulary remains useful when the device and event-construction interface change.

Cross-dataset conclusions also depend on how transfer is evaluated. A representation can achieve low out-of-domain perplexity because the target stream is easier to predict, while a small or nearly deterministic vocabulary can appear stable despite retaining little motion information. We therefore combine degradation with target-domain regret and low-order target references, paired bootstrap, order sensitivity, motif overlap, and frozen structural probes. We further stress-test the conclusions across tokenizer configurations, temporal scales, VQ codebook policies, and alternative event constructions. This evaluation jointly examines predictive transfer, target predictability, and structural retention.

The results are quite direction-specific. Angular-motion tokens have lower target-domain regret than frozen-codebook VQ tokens for AEA$\rightarrow$RITW, while RITW$\rightarrow$AEA is inconclusive under paired bootstrap. Angular-motion tokens nevertheless retain stronger order dependence and motif overlap, while the controlled probes show complementary representation properties. On EGTEA, a matched three-interface comparison shows that event construction materially changes transfer: native events have the lowest regret into EGTEA, degree-aware I-VT retains substantial motif overlap, and frame-span events have zero motif overlap and severe EGTEA-as-source transfer. These findings show that motion-token transfer depends not only on the vocabulary, but also on the temporal and event interfaces through which gaze is represented.

Our contributions are fourfold. First, we formulate event-aligned, fixed-horizon angular displacement as an interpretable, event-conditioned motion vocabulary and compare it with discrete, learned, and continuous alternatives under cross-dataset transfer. Second, we provide an evaluation protocol that contextualizes transfer using target references and tests whether temporal and geometric structure remain available. Third, we analyze tokenizer sensitivity, temporal granularity, codebook policy, and representation-specific failure modes. Fourth, we compare degree-aware I-VT, vendor-native, and frame-span constructions on a third egocentric dataset under matched training and evaluation controls, showing that event construction is part of the representation rather than neutral preprocessing.

\section{Related work}

\paragraph{Gaze representation.}
The choice of gaze representation is not neutral, as different formulations preserve different aspects of behavior while discarding others. Classical saccade analysis characterizes movements through angular amplitude, duration, and velocity~\cite{bahill1975main}, while ScanMatch~\cite{cristino2010scanmatch} converts fixations into spatial-temporal letter sequences, and MultiMatch~\cite{dewhurst2012depends} compares scanpaths across shape, position, direction, length, and duration. Other sequence-based methods represent gaze through latent-state transitions~\cite{coutrot2018scanpath,chuk2014understanding}, recurrence patterns~\cite{anderson2013recurrence}, or short subsequence frequencies~\cite{kubler2017subsmatch}. These methods differ in what structure they make measurable, reinforcing that representation choice is part of the modeling problem.

\paragraph{Discrete gaze tokenization.}
Discrete scanpath representations are common in comparison and prediction: they make alignment, sequence modeling, and motif analysis tractable, but trade off fidelity for stability~\cite{anderson2015comparison,laborde2026vision}. Recent work explicitly compares tokenization strategies for gaze forecasting and generation with language-model-style architectures~\cite{rolff2025tokenization}. Our focus takes a complementary evaluation angle. Beyond asking whether gaze should be continuous or discrete, we ask which abstraction preserves reusable sequential structure under dataset shift and how conclusions change with target-domain predictability, codebook construction, and event definition. To this end, we treat angular-delta tokenization as a representation family and test its robustness to grid size, stride, event conditioning, zero-motion handling, and out-of-range handling.

\paragraph{Domain shift and transfer.}
Domain shift is well known in gaze estimation, where models degrade across datasets because capture conditions, subjects, gaze ranges, and annotation procedures change~\cite{zhang2015appearance,kellnhofer2019gaze360,zhang2017mpiigaze}. Higher-level gaze models also increasingly use structured gaze or person-specific tokens to improve generalization~\cite{tafasca2024sharingan,gupta2024mtgs,li2026thinking}. Most work measures robustness through end-task accuracy. Instead, we ask whether the representation itself remains useful under shift: compact enough to be learned efficiently, structured enough to support prediction, and stable enough to preserve local sequential structure across datasets.


\section{Method and Experimental Setup}

\rev{We study cross-domain gaze sequence modeling to determine which representations remain useful under dataset shift. Given a source dataset, we tokenize gaze events, train a next-token model on the source, and evaluate it both in-domain and on held-out target-domain streams. Here, transfer refers to predictive and structural retention under a specified source–target protocol. It is not assumed to hold across incompatible event, coordinate, temporal, or task interfaces.}

\subsection{Datasets}

We use three egocentric eye-tracking datasets with distinct behavioral regimes: AEA (general everyday activities)~\cite{lv2024aria}, RITW (Reading in the Wild--Columbus dataset)~\cite{yang2025reading}, and EGTEA (cooking activities)~\cite{li2021eye}. AEA and RITW were both recorded with Meta's Project Aria Gen~1 headset~\cite{engel2023project}, while EGTEA was recorded with an SMI wearable eye tracker~\cite{li2021eye}. \rev{In the gaze exports used in our pipeline, AEA and RITW differ substantially in effective sampling density.} EGTEA differs in both sensor and export format, as it is provided through BeGaze exports with frame-aligned gaze samples and native fixation/saccade annotations. Consequently, AEA and RITW are used as the main cross-dataset transfer pair and EGTEA as an event-interface boundary check.

\subsection{Representations and models}
\label{sec:reps}

All discrete representations are derived from event streams. For AEA and RITW, we obtain events using a simplified adaptive velocity-threshold identification (I-VT) procedure inspired by~\citeauthor{nystrom2010adaptive}~\cite{nystrom2010adaptive}, with a median absolute deviation (MAD)-based per-sequence velocity threshold. For EGTEA, we compare three event interfaces. Degree-aware I-VT constructs fixation/saccade events after applying $360^\circ$ angular unwrapping to the projection-based pseudo-yaw/pitch values. The
native interface uses the BeGaze fixation/saccade annotations and averages valid samples within each interval. The frame-span interface emits one fixation-typed event per retained video frame within contiguous valid spans. All three interfaces use the same AD tokenizer. Full construction and health statistics are provided in the supplement.

Our primary representation is \texttt{ang\_delta} (AD), a deterministic motion tokenizer that quantizes event-conditioned $(\Delta yaw,\Delta pitch)$ on a fixed 2D grid, with explicit zero-motion tokens and configurable out-of-range handling. Rather than differencing adjacent event centers, the default AD interpolates yaw/pitch $80$\,ms after each valid event, subject to the stride tolerance, and quantizes the resulting displacement. If no valid target exists, it emits the event-specific zero token. We compare AD with \texttt{anchor\_ang} (AA), discretized absolute yaw/pitch; \texttt{ang\_state\_and\_motion} (ASM), interleaved absolute state and local motion; \texttt{so3\_delta} (SD), the $\mathrm{SO}(3)$ log-map between consecutive gaze rays; \texttt{event\_only} (EVT), an event-label baseline; \texttt{spatial\_grid+event} (SGE), event type combined with a fixed yaw/pitch cell; and \texttt{vq\_delta} (VQD), which maps normalized $(\Delta yaw,\Delta pitch)$ to the nearest frozen-codebook centroid. For event-conditioned tokenizers, the event label and representation bin jointly index one token. The Transformer uses a single embedding lookup without a separate event-type embedding. %
For VQD, we evaluate $K=128$ codebooks over three seeds under source-specific, AEA-reuse, and pooled AEA+RITW fitting. All codebooks use training deltas only. A fixed-seed $K=256$ result provides a secondary size diagnostic. For the controlled AD--VQD regret comparison, the source-fitted $K=128$ codebook remains frozen during source-model training, target-reference training, and target evaluation, ensuring a fixed discrete alphabet.

All discrete representations share the same next-token Transformer architecture, splits, context length, minimum sequence-length filter, optimizer family, and training budget. Only the AD temporal-scale analysis in Section~\ref{sec:temporal_scale_analysis} varies stride and context length. The supplement details each tokenizer. %
We also include a continuous autoregressive baseline (CONT), which predicts normalized continuous $\Delta yaw$ and $\Delta pitch$ using a Transformer trained with Gaussian negative log-likelihood. Because it belongs to a different metric family, CONT is reported separately. Model and training details appear in the supplement.

\subsection{Evaluation protocol}

For discrete models, we accumulate token negative log-likelihood in nats. We report next-token perplexity and convert target cross-entropy to bits/token for regret:
\begin{equation}
\begin{aligned}
H_T^{\mathrm{nat}}(q)
    &= -\frac{1}{N}\sum_{t=1}^{N}\ln q(x_t\mid x_{<t}),\\
H_T(q)
    &= \frac{H_T^{\mathrm{nat}}(q)}{\ln 2}, \qquad
\mathrm{PPL}(q)=\exp\!\left(H_T^{\mathrm{nat}}(q)\right).
\end{aligned}
\end{equation}
{
\begin{equation}
\mathrm{Degradation}=
\frac{\mathrm{Metric}_{\mathrm{OOD}}-\mathrm{Metric}_{\mathrm{IND}}}
     {\mathrm{Metric}_{\mathrm{IND}}}.
\end{equation}
}
Lower perplexity is better, while smaller degradation indicates greater stability under shift. Because degradation neither controls for target predictability nor provides statistical separation, and may be negative when the target stream is easier, we interpret it alongside regret and structural diagnostics. \rev{Likewise, alphabets and token rates differ---ASM emits two tokens per event--- so perplexity and bits/token are representation-specific, and equal contexts may cover different event histories. They diagnose transfer, not motion fidelity.}

Target-domain regret partially controls for target predictability by comparing source- and target-trained models. Let $H_T$ be cross-entropy on the held-out target test split, $q_S$ a source-trained model, $q_T$ the same architecture trained on the target training split, and $q_B$ a target-trained unigram or first-order Markov baseline. All are evaluated on the same target test split.
{
\begin{equation}
\begin{aligned}
R_{\mathrm{raw}}(S\!\rightarrow\!T)
    &= H_T(q_S)-H_T(q_T),\\
R_{\mathrm{norm}}^{(B)}(S\!\rightarrow\!T)
    &= \frac{H_T(q_S)-H_T(q_T)}
            {H_T(q_B)-H_T(q_T)}.
\end{aligned}
\end{equation}
}
Normalized regret above one means that source-trained transfer remains worse than baseline $B$ relative to the target-trained reference. A non-positive denominator makes normalized regret undefined and is reported as NA. For AD--VQD comparisons, we derive matched differences from sequence-level losses weighted by valid prediction counts and obtain 95\% confidence intervals using a hierarchical bootstrap over seeds and target sequences.

To verify that models capture sequence structure beyond merely token frequencies, we use order-manipulation controls based on perplexity: RR (train real, test real), RS (train real, test shuffled), and SR (train shuffled, test real). Using these controls, we define:
\[
\begin{aligned}
\mathrm{Order\ Sensitivity}
    &= \frac{\mathrm{RS}-\mathrm{RR}}{\mathrm{RR}},\\
\mathrm{Learning\ Gap}
    &= \frac{\mathrm{SR}-\mathrm{RR}}{\mathrm{RR}}.
\end{aligned}
\]

Order sensitivity measures the effect of disrupting test order, while learning gap measures the cost of training on shuffled sequences. We additionally measure local cross-domain compatibility using bigram and trigram Jaccard overlap and the target $n$-gram mass unseen in the source. Counting and top-$K$ details are given in the supplement.

As structural diagnostics, we train linear probes on frozen, mean-pooled sequence-model states to predict motion-state, motion-quadrant, and direction-reversal decodability. These assess local motion-state, motion-quadrant, and reversal information. The supplement provides full probe details and a secondary within-dataset RITW reading-state probe.

We analyze AD temporal scale by sweeping event-pair stride (\texttt{delta\_stride\_ms}) and context length (\texttt{context\_len}) in both transfer directions. Stride controls motion per token, while context length controls the available token history. We report OOD perplexity and degradation and examine token-distribution statistics and Kneser--Ney-smoothed $n$-gram predictability~\cite{kneser1995improved,chen1999empirical} to distinguish \textit{compression} from \textit{aggregation} effects.

We also test AD sensitivity to angular-grid resolution, event-pair stride, and event-type conditioning over multiple seeds. Each variant uses the full AEA$\leftrightarrow$RITW protocol and reports IND/OOD perplexity, degradation, order sensitivity, entropy, top-10 concentration, and vocabulary usage. Fixed-seed checks cover zero-motion and out-of-range handling.

For EGTEA, we compare degree-aware I-VT, vendor-native, and frame-span interfaces. All conditions use fixed context 8, minimum sequence length 20, three seeds, identical session-level 70/15/15 splits, uniform session sampling, and matched per-session window caps. Transfer degradation, target-domain regret, motif overlap, and source-mismatch statistics are evaluated on the same 86 sessions.

\subsection{Reproducibility}

We repeat the main AEA$\leftrightarrow$RITW comparison, tokenizer-sensitivity settings, continuous baseline, and structural-probe diagnostics over three seeds and report mean and standard deviation. Deterministic tokenizers have fixed token streams and varying model seeds. VQD uses source-specific $K=128$ codebook seeds, and CONT is trained separately per seed. The matched EGTEA event-interface comparison uses three seeds and population standard deviations. The $\mathrm{SO}(3)$ ablations remain fixed-seed seed-0 diagnostics. Code and settings are available at \url{https://anonymous.4open.science/r/motionTokenizer_egocentric-BE1E/}.


\section{Results}
\subsection{Cross-domain transfer}

\begin{figure*}[t]
    \centering
    \includegraphics[width=1\textwidth]{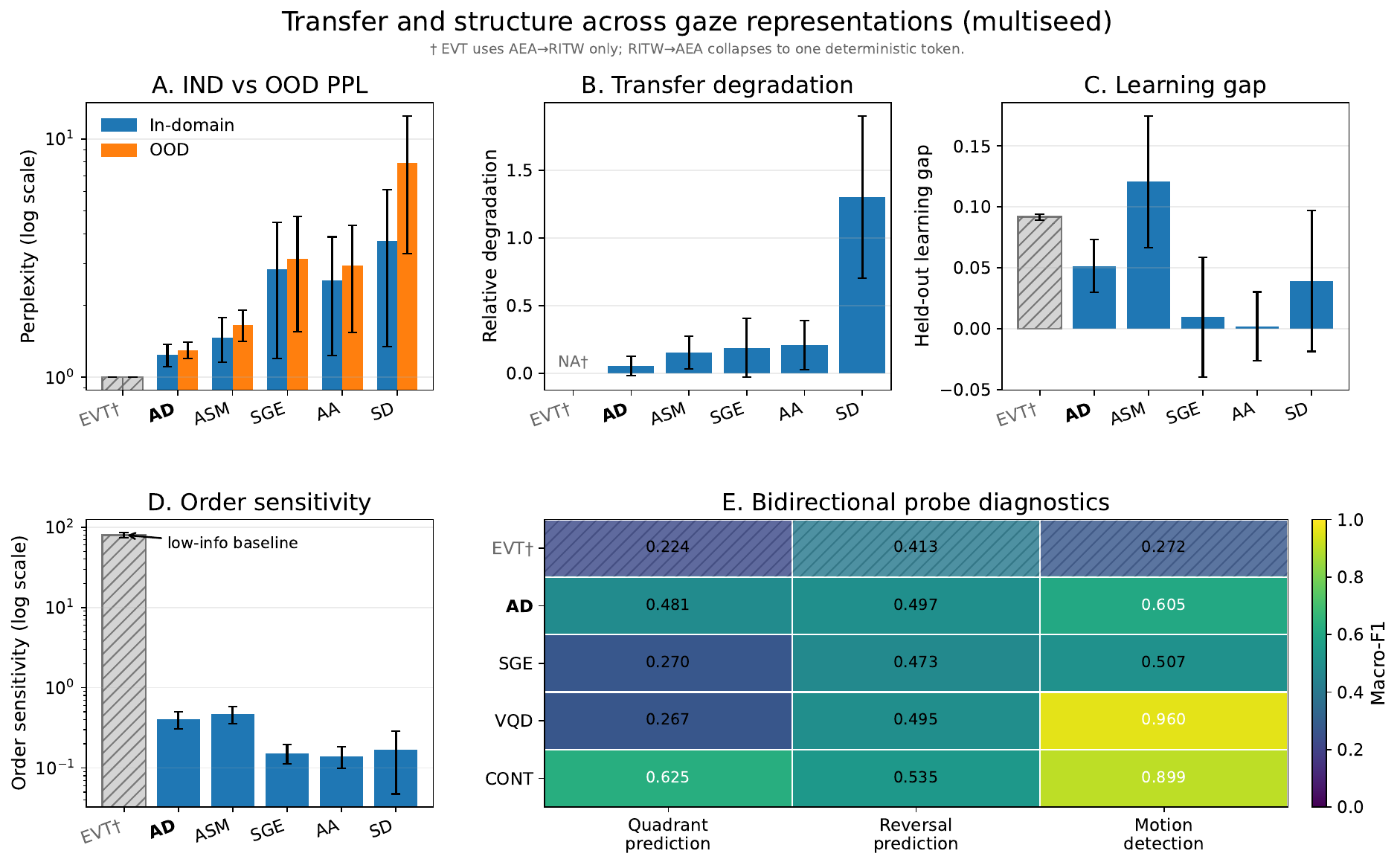}
   \caption{Transfer and structure across gaze representations on the bidirectional AEA$\leftrightarrow$RITW benchmark. Panels A--D summarize the hand-designed tokenizer comparison. Panel E shows the matched context-16 probe rerun with source-specific $K=128$ VQD and CONT. EVT$^\dagger$ uses AEA$\rightarrow$RITW only in Panels A, C, and D. Panel B is NA because the RITW$\rightarrow$AEA stream collapses to one deterministic token, making bidirectional degradation undefined. Abbreviations follow Sec.~\ref{sec:reps}.}
    \label{fig:main_transfer}
\end{figure*}

Our main discrete benchmark is AEA$\leftrightarrow$RITW (see Figure~\ref{fig:main_transfer} for a visualization of the transfer results). Excluding the geometry-free EVT control, AD has the lowest observed pooled mean degradation among the hand-designed tokenizers. It obtains degradation $0.05 \pm 0.07$, lower than ASM ($0.15 \pm 0.12$), SGE ($0.19 \pm 0.22$), AA ($0.21 \pm 0.18$), and SD ($1.30 \pm 0.60$). Direction-specific degradation is $0.01 \pm 0.08$ for AEA$\rightarrow$RITW and $0.09 \pm 0.04$ for RITW$\rightarrow$AEA. \rev{Because degradation does not control for target-domain predictability, we next compare AD and VQD using target-domain regret.}

\begin{table*}[t]
\centering
{
\caption{Target-domain regret for AD and VQD under the controlled context-16 AEA$\leftrightarrow$RITW comparison. $H_T(q_S)$ and $H_T(q_T)$ are target-test cross-entropies in bits/token for source- and target-trained models. Lower raw regret is better. Normalized values above one indicate worse performance than the corresponding target low-order baseline. Values are means over seeds. VQD normalization for RITW$\rightarrow$AEA is NA because the denominator is non-positive.}
\label{tab:target_regret}
\scriptsize
\setlength{\tabcolsep}{4pt}
\begin{tabular}{llccccc}
\toprule
Transfer & Rep. & $H_T(q_S)$ & $H_T(q_T)$ &
Raw regret $\downarrow$ & Unigram norm. $\downarrow$ &
Markov norm. $\downarrow$ \\
\midrule
AEA$\rightarrow$RITW & AD  & 2.924 & 2.290 & 0.634 & 0.450 & 1.958 \\
AEA$\rightarrow$RITW & VQD & 5.291 & 3.672 & 1.620 & 2.516 & 3.889 \\
\midrule
RITW$\rightarrow$AEA & AD  & 3.529 & 3.026 & 0.503 & 0.373 & 0.265 \\
RITW$\rightarrow$AEA & VQD & 7.342 & 6.814 & 0.528 & NA & NA \\
\bottomrule
\end{tabular}
\vspace{2pt}
\begin{minipage}{0.97\textwidth}
\end{minipage}
}
\end{table*}

\rev{Because Table~\ref{tab:target_regret} uses a padded single-target context-16 protocol, its cross-entropies are not comparable with Figure~\ref{fig:main_transfer}A and support only within-protocol regret comparisons. In AEA$\rightarrow$RITW, AD has lower raw target-domain regret than VQD (0.634 vs.\ 1.620 bits/token), and VQD remains worse than both target-domain low-order baselines after normalization. In RITW$\rightarrow$AEA, raw regrets are close (0.503 for AD and 0.528 for VQD), and VQD normalized regret is undefined because the low-order baseline denominator is non-positive.}


\rev{The paired AD--VQD raw-regret difference is $-0.964$ bits/token for AEA$\rightarrow$RITW, with 95\% CI $[-1.183,-0.756]$, confirming lower AD regret. For RITW$\rightarrow$AEA, the difference is $0.066$, with 95\% CI $[-0.417,0.616]$, making this direction inconclusive. Because these estimates use matched sequence-level losses, they need not equal differences between the seed-level means in Table~\ref{tab:target_regret}. Full bootstrap results are reported in the supplement.}

EVT shows why perplexity is not enough on its own. Its near-perfect AEA$\rightarrow$RITW transfer follows from a tiny, highly predictable vocabulary, but it contains no geometric or motion information. In the opposite direction, the metric becomes degenerate. Its high order sensitivity reflects an almost deterministic event sequence that is fragile when order is disrupted. EVT is thus used only as a low-information sanity check. This illustrates a general failure mode for token-level transfer metrics: a representation can appear robust because it has collapsed the behavioral signal into a small, nearly deterministic alphabet. For this reason, we interpret low degradation only together with structural diagnostics that test whether geometry and temporal order remain available.

SD shows that preserving more geometric detail is not automatically beneficial. The default full angle-axis $\mathrm{SO}(3)$ tokenizer transfers poorly in both directions and has almost no shared local motif structure, with bigram/trigram overlap of only 0.054/0.002. This failure is not mainly due to outside buckets or missing individual tokens, given that outside clipping changes little and unigram support overlap remains moderate. Instead, the ablation suggests axis-conditioned fragmentation. Collapsing SD to an angle-only variant recovers much stronger transfer and motif overlap, raising bigram/trigram overlap to 0.683/0.418. SD thus serves as a useful failure case, showing how fine-grained 3D rotation tokens can over-partition local gaze motion, making transition structure less reusable across datasets. Further $\mathrm{SO}(3)$ ablations are reported in the supplement.

\subsection{Representation diagnostics}


\paragraph{Order sensitivity and learning gap.}
Figures~\ref{fig:main_transfer}C--D show that AD depends meaningfully on real sequence order. Across three seeds, AD has pooled order sensitivity $0.40 \pm 0.10$, lower than ASM ($0.47 \pm 0.11$) but substantially higher than AA ($0.14 \pm 0.04$), SGE ($0.15 \pm 0.04$), and SD ($0.17 \pm 0.12$). This pattern suggests that both AD and ASM capture sequential structure.

The learning-gap control gives a compatible picture. AD has a moderate pooled learning gap ($0.05 \pm 0.02$), while ASM is higher ($0.12 \pm 0.05$). AA, SGE, and SD remain lower or noisier. Thus, AD combines relatively low degradation with substantial order sensitivity. Motif overlap further separates the representations: AD has much higher bigram/trigram Jaccard overlap ($0.493/0.449$) than ASM ($0.173/0.146$), AA ($0.031/0.003$), SGE ($0.042/0.003$), or SD ($0.034/0.005$). This supports the interpretation that AD preserves reusable local transition structure \rev{under the shared AEA/RITW event and coordinate interface.}

\paragraph{Probe-based diagnostics.}
\rev{Figure~\ref{fig:main_transfer}E summarizes the matched context-16 bidirectional frozen-probe rerun using the source-specific $K=128$ VQD policy. Reported means and standard deviations pool both transfer directions over three seeds, giving six direction-by-seed values. The probes show complementary strengths. CONT has the highest mean motion-quadrant decodability ($0.625 \pm 0.101$), while AD is highest among the discrete tokenizers ($0.481 \pm 0.044$), compared with EVT ($0.224 \pm 0.012$), SGE ($0.270 \pm 0.028$), and VQD ($0.267 \pm 0.025$). Direction-reversal decodability is less separated: CONT reaches $0.535 \pm 0.044$, while AD, VQD, SGE, and EVT reach $0.497 \pm 0.044$, $0.495 \pm 0.026$, $0.473 \pm 0.031$, and $0.413 \pm 0.029$, respectively.}

\rev{Motion detection instead measures generic movement decodability. VQD and CONT have the highest means ($0.960 \pm 0.030$ and $0.899 \pm 0.091$), while AD reaches $0.605 \pm 0.103$. Thus, VQD and CONT make generic motion information more accessible, whereas AD has higher motion-quadrant decodability than the other discrete tokenizers. Because the representations differ in temporal pairing, these probes assess local-motion decodability rather than provide a controlled comparison of future-motion prediction. They measure structural retention, not cross-dataset video-understanding utility. The within-dataset reading-state probe remains a secondary diagnostic in the supplement.}

\subsection{Tokenizer sensitivity}
\label{sec:tokenizer_sensitivity}

\rev{The default AD setting is treated as a fixed reference as opposed to an optimized endpoint. The full multiseed audit is reported in the supplement.} Grid size shows a collapse--fragmentation trade-off. A coarse $12{\times}9$ grid has low degradation ($0.03 \pm 0.04$) and high order sensitivity, but concentrates over $0.96$ of AEA and $0.98$ of RITW token mass in the top 10 tokens. A finer $32{\times}24$ grid uses more vocabulary but transfers worse than the default and has lower order sensitivity. Removing event type lowers degradation ($0.02 \pm 0.07$) but sharply reduces order sensitivity ($0.06 \pm 0.04$), so event conditioning is retained. At the default context length, the $40$\,ms stride has the highest variance and worst mean degradation, while $120$\,ms remains competitive but does not dominate the default. This does not contradict the temporal sweep below, where $40$\,ms becomes useful for RITW$\rightarrow$AEA only with a longer context.

\subsection{Temporal scale analysis}
\label{sec:temporal_scale_analysis}

The best temporal setting depends on transfer direction. For AEA$\rightarrow$RITW, a stride of $120$\,ms with context length $32$ gives the lowest OOD perplexity ($1.21 \pm 0.07$). For RITW$\rightarrow$AEA, the best setting uses a $40$\,ms stride and context length $64$ ($1.022 \pm 0.004$). There is therefore no single best temporal scale. 

Equal effective temporal horizon does not imply equal transfer. At a $2560$\,ms horizon, AEA$\rightarrow$RITW obtains OOD perplexity $1.24 \pm 0.11$ with $(80\,\mathrm{ms},32)$ but $5.27 \pm 1.07$ with $(40\,\mathrm{ms},64)$. In the reverse direction, $(40\,\mathrm{ms},64)$ gives $1.022 \pm 0.004$, compared with $1.24 \pm 0.03$ for $(160\,\mathrm{ms},16)$. Token granularity and context length therefore contribute independently.

Follow-up diagnostics suggest two mechanisms. Moderate stride-level smoothing improves the AEA token distribution, increasing entropy from $2.43$ at $40$\,ms to $3.29$ at $120$\,ms while reducing top-10 concentration from $0.94$ to $0.72$. Smoothed $n$-gram analysis shows modest gains beyond the bigram level for RITW, whereas AEA saturates earlier. These results are consistent with respective compression and aggregation effects, making the optimal combination direction-dependent. The full heatmap and matched-horizon comparisons are reported in the supplement.


\subsection{Comparison to continuous and VQ baselines}

The continuous autoregressive baseline is evaluated in a different metric family from the discrete token models. Across three seeds, AEA$\rightarrow$RITW gives IND MSE $0.64 \pm 0.11$ and OOD MSE $0.16 \pm 0.04$, with MSE degradation $-0.74 \pm 0.07$. However, the corresponding OOD $R^2$ is close to zero and slightly negative ($-0.04 \pm 0.17$), despite positive IND $R^2$ ($0.36 \pm 0.02$), consistent with a change in target-domain variance and scale. In the reverse direction, RITW$\rightarrow$AEA is much harder: OOD MSE rises to $5.37 \pm 1.46$ and MSE degradation reaches $7.59 \pm 2.57$, while OOD $R^2$ remains positive but lower than IND $R^2$ ($0.18 \pm 0.04$ vs.\ $0.37 \pm 0.13$). These results still make the continuous model a useful predictive baseline, but not a direct replacement for the discrete transfer table, because it is evaluated with continuous regression metrics and does not expose token-level transition structure.

The VQD baseline is viable but codebook-policy sensitive. Our main VQD setting uses a source-specific $K=128$ codebook, which has the lowest mean degradation over AEA$\leftrightarrow$RITW among the tested VQ policies ($-0.07 \pm 0.10$). The AEA-reuse policy remains competitive ($-0.04 \pm 0.14$), while the pooled codebook is weaker in transfer ($0.03 \pm 0.02$) and almost eliminates motif overlap despite using nearly all centroids. Its bigram/trigram Jaccard values fall to $0.005 \pm 0.003$ and $0.003 \pm 0.000$, compared with $0.227 \pm 0.022$ and $0.034 \pm 0.011$ for AEA reuse. Thus, VQ codebook policy can reduce distributional mismatch without preserving reusable local token structure. Source-specific fitting has the lowest mean degradation, whereas AEA reuse preserves stronger motif overlap. This reinforces that VQ transfer performance and local-structure preservation are not the same criterion. We treat VQD as a useful learned-motion baseline, but its structural interpretation depends on codebook provenance. The full codebook-policy results are reported in the supplement.

\subsection{EGTEA: event construction as a representation boundary}

EGTEA tests whether motion-token structure survives changes in device, export format, and event construction. We compare degree-aware I-VT, vendor-native fixation/saccade events, and frame-span events under fixed context 8, minimum sequence length 20, three seeds, identical session splits, uniform session sampling, and matched per-session window caps. All interfaces retain the same 86 sessions.

The resulting streams differ substantially. I-VT retains 61{,}587 events with median sequence length 571 and entropy 5.03 bits. Native events retain 473{,}683 events with median length 4{,}090.5 and entropy 3.97 bits. Frame-span construction emits 1{,}125{,}577 fixation-typed frame events, but its sequences are much shorter (median 35), more concentrated (top-10 share 0.93), and contain more zero-motion tokens (0.065).

Native events have the lowest target-domain regret for transfer into EGTEA: $0.172\pm0.021$ bits/token from AEA and $0.297\pm0.036$ from RITW. The corresponding I-VT regrets are
$0.578\pm0.014$ and $1.268\pm0.045$, while frame-span regrets are $1.085\pm0.017$ and $1.209\pm0.071$. When EGTEA is the source, native-event regret is $0.085\pm0.017$ into AEA and
$-0.015\pm0.010$ into RITW. I-VT gives $-0.149\pm0.061$ and $-0.024\pm0.008$, whereas frame-span regret rises to $5.806\pm0.129$ and $6.048\pm0.125$. Negative regret means
that the source-trained model scores below the finite target-trained reference under this protocol. It should not be interpreted as outperforming an oracle.

The motif diagnostics show a compatible distinction. I-VT bigram/trigram Jaccard@50 is $0.648/0.381$ with AEA and $0.641/0.531$ with RITW. Native-event overlap is $0.464/0.276$ with
AEA and $0.746/0.746$ with RITW. Frame-span overlap is zero for both datasets, and all target bigram and trigram mass is unseen across the frame-span boundary.

These results isolate the downstream protocol but not the defining properties of each interface. The interfaces necessarily differ in event boundaries, temporal units, labels, sequence lengths, and event conditioning. The conclusion is therefore not that one interface is universally preferable, but that the upstream definition of an event
materially changes the structure and transfer behavior exposed to the same tokenizer.

\section{Discussion}

\paragraph{Representation choice determines what transfers.}
The main result of this paper is that the choice of discretization determines which structure survives transfer. \rev{In the AEA$\leftrightarrow$RITW benchmark, angular motion tokens provide a useful hand-designed compromise. They are more informative than event labels, less tied to absolute spatial state, and compact enough for reuse across related egocentric datasets. Against VQD, AD has lower target-domain regret for AEA$\rightarrow$RITW, while the reverse direction is inconclusive.} EVT shows why low perplexity alone can be misleading: its near-deterministic stream is easy to predict but geometrically empty. \rev{In this compatible setting, AD captures local motion with reduced dependence on absolute gaze position under a shared event and coordinate interface, while remaining coarse enough to avoid severe token-space fragmentation.}

\paragraph{Expressiveness does not guarantee transfer.}
The alternative motion baselines illustrate how different representation choices affect transfer. The evaluated $\mathrm{SO}(3)$ tokenizer is geometrically expressive, but jointly quantizing rotation angle and axis direction fragments local motion structure---bigram and trigram support nearly vanishes across AEA/RITW, and the angle-only ablation recovers much stronger transfer. VQD shows a different alignment problem: the learned codebook must partition motion space in a way that remains structurally meaningful across domains. A pooled codebook can use centroids efficiently while destroying motif overlap, so codebook usage alone is not enough. In both cases, predictive fit and structural reuse diverge: SD fragments the motion vocabulary despite its geometric expressiveness, while VQD can smooth transfer without preserving reusable local transitions. These comparisons evaluate complete representations: AD and VQD differ in temporal pairing and event conditioning as well as quantization, so they do not isolate the quantizer alone.

\paragraph{Temporal scale is part of the representation.}
The temporal sweep shows that token scale is not just a downstream model hyperparameter. Equal effective temporal horizon does not imply equal transfer: AEA$\rightarrow$RITW benefits from moderate stride-level smoothing, while RITW$\rightarrow$AEA prefers finer tokens with longer context. Mechanism analyses suggest a compression effect for AEA, where moderate stride increases token entropy and reduces concentration, and a weaker aggregation effect for RITW, where smoothed $n$-gram analysis indicates slightly more usable longer-context structure. We do not test scene context causally, so the safest conclusion is that different gaze regimes may require different temporal compression before their structure becomes predictively useful under transfer.

\paragraph{Event interfaces are part of the representation.}
The matched EGTEA comparison strengthens the event-interface result. Native events have the lowest target-domain regret for transfer into EGTEA, while degree-aware I-VT retains substantial motif overlap but shows higher regret into EGTEA. Frame-span events have zero local motif overlap and fail severely when EGTEA is used as the source.
Because context, splits, optimizer updates, session sampling, and window caps are controlled, these differences cannot be attributed to the earlier protocol mismatch. They still reflect the complete interfaces, which differ intrinsically in boundaries, temporal units, labels, sequence lengths, and event conditioning. Sharing a motion vocabulary is therefore insufficient. The upstream interface determines which transitions the tokenizer receives.

\paragraph{Practical implications.}
Motion-based gaze tokens provide a compact input for temporal-attention models, but tokenizer choice remains part of model design. Token streams should be checked for collapse, order sensitivity, and cross-domain motif mismatch, because poor transfer may reflect either a different gaze regime or an incompatible event interface.


\paragraph{Limitations and future work.}
\rev{The study covers three egocentric datasets, so its conclusions concern event-based egocentric gaze transfer under the tested interfaces rather than gaze representation generally.} The clearest comparative evidence comes from AEA and RITW, which share the Project Aria device family. Their common sensor geometry, coordinate conventions, and preprocessing assumptions likely make transfer easier than across arbitrary eye trackers or stimulus regimes. The datasets also differ in sequence length and available training windows, which can interact with context length and order-manipulation controls. We address this through matched protocols and temporal-scale analyses, but larger balanced egocentric corpora are needed to separate dataset size, task regime, and sensor effects more cleanly. The continuous and discrete baselines use different metric families and should therefore be interpreted as complementary rather than directly interchangeable. \rev{Finally, the supervised probes are lightweight diagnostics of structural retention and behavioral decodability, not evidence of cross-dataset video-understanding utility. Stronger application-level tests should combine gaze tokens with egocentric video for activity recognition, task-state prediction, gaze-conditioned video understanding, or attention-guidance systems.}

\paragraph{Privacy and Ethical Considerations.}
Discrete motion tokens should not be treated as anonymization. They remove some spatial precision and may make exact coordinate- or stimulus-level reconstruction harder, but they intentionally preserve temporal structure. That structure can still encode task, context, user state, ability, or interaction patterns. Tokenization may therefore reduce some risks of raw gaze traces while transforming others into sequence-level behavioral signatures.

\section{Conclusion}

This paper studies gaze tokenization as a representation problem and asks which abstraction remains useful for egocentric event-based transfer under dataset shift. \rev{In the AEA$\leftrightarrow$RITW benchmark, AD has lower target-domain regret than VQD for AEA$\rightarrow$RITW, while RITW$\rightarrow$AEA remains inconclusive under paired bootstrap. AD preserves useful order sensitivity and motif overlap under the shared event and coordinate interface, whereas continuous and VQD representations retain complementary motion structure in the probes.} Temporal and tokenizer-sensitivity analyses show that token granularity, context length, grid design, event conditioning, and zero/out-of-range handling all shape what structure becomes learnable. A matched EGTEA comparison further shows that degree-aware I-VT, vendor-native, and frame-span interfaces expose markedly different
transfer and motif structure to the same tokenizer, with frame-span events failing severely when used as the source. \rev{Motion-based tokenization is therefore a compact and interpretable option for compatible egocentric gaze streams, provided that tokenizer design, event interfaces, and retained behavioral structure are explicitly considered.} The broader lesson is that cross-dataset gaze modeling requires aligning not only models, but also the behavioral units and token vocabularies through which gaze is exposed to those models.

{
    \small
    \bibliographystyle{ieeenat_fullname}
    \bibliography{main}
}
\end{document}